\documentclass{article} % For LaTeX2e
\usepackage{iclr2027_conference,times}

\usepackage{amsmath,amsfonts,bm}

\def\eqref#1{equation~\ref{#1}}
\def\1{\bm{1}}

\DeclareMathAlphabet{\mathsfit}{\encodingdefault}{\sfdefault}{m}{sl}
\SetMathAlphabet{\mathsfit}{bold}{\encodingdefault}{\sfdefault}{bx}{n}

\usepackage{hyperref}
\usepackage{url}
\usepackage{fontawesome5}
\usepackage{xcolor}
\usepackage{hyperref}
\usepackage{graphicx}
\usepackage{tabularx}
\usepackage{booktabs}
\usepackage{xspace}
\usepackage{listings}
\usepackage{cleveref}
\usepackage[most]{tcolorbox}
\usepackage{enumitem}
\newtcolorbox{promptbox}[1][]{
  breakable,
  colback=gray!5,
  colframe=gray!60,
  fonttitle=\bfseries,
  title=#1,
  left=2mm, right=2mm, top=1mm, bottom=1mm,
  fontupper=\small,
  before upper={\setlength{\parskip}{0.5em}}
}

\definecolor{exampleblue}{HTML}{227BB5}

\newtcolorbox{examplebox}[1]{
  enhanced,
  colback=exampleblue!5!white,
  colframe=exampleblue,
  colbacktitle=exampleblue,
  coltitle=white,
  fonttitle=\bfseries,
  title={#1},
  halign title=center,
  boxrule=1pt,
  arc=3mm,
  left=4mm,
  right=4mm,
  top=3mm,
  bottom=3mm,
  before skip=8pt,
  after skip=8pt
}

\newcommand{\ph}[1]{\textcolor{blue!60!black}{\texttt{\{#1\}}}}
\newcommand{\ourbench}{PlotGround-1k\xspace}

\title{PlotGround: Grounding Plot Digitization in Real Scientific Figures and Their Source Data}

\author{%
  Yaohui Zhang$^{1,3}$\thanks{Co-first authors} \quad
  Binxu Li$^{1,2}$\footnotemark[1] \quad
  Haoyi Duan$^{2}$\footnotemark[1] \quad
  Jiacheng Miao$^{3}$ \quad
  \textbf{Yixin Wang}$^{3}$ \quad
  \\
  \textbf{Xinran Du}$^{2}$ \quad
  \textbf{Chenyue Li}$^{3}$ \quad
  \textbf{Shilong Liu}$^{2}$ \quad
  \textbf{Kevin Wu}$^{1}$ \quad
  \textbf{James Zou}$^{1,3}$
  \\
  $^{1}$Generative Expert Labs, Inc \quad
  $^{2}$Princeton University \quad
  $^{3}$Stanford University
\\[0.5em]
\makebox[\linewidth][c]{%
  \href{https://www.staging.gxl.ai/blog/plotground/}{%
    \textcolor{blue!50!black}{%
      \faGlobe\;\textbf{PlotGround Website}%
    }%
  }%
  \qquad
  \href{https://huggingface.co/datasets/yhzhang3/PlotGround-1k}{%
    \raisebox{-0.2em}{%
      \includegraphics[height=1.2em]{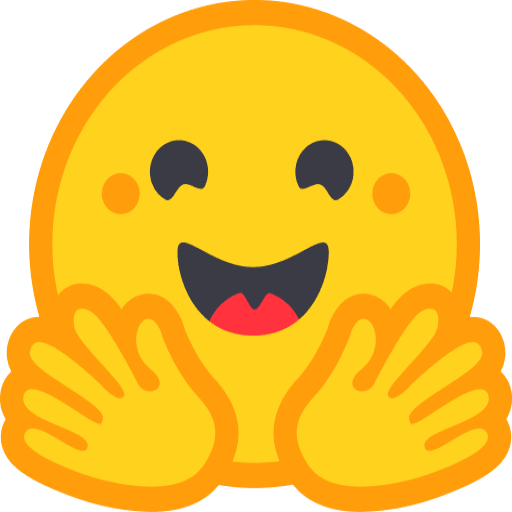}%
    }%
    \;\textcolor{blue!50!black}{\textbf{Hugging Face Dataset}}%
  }%
}
}

\iclrfinalcopy % Uncomment for camera-ready version, but NOT for submission.
\begin{document}

\maketitle

\begin{abstract}
Scientific figures often encode quantitative results that are not readily available in machine-readable form, making accurate plot digitization important for verifying and reusing published findings. Yet it remains unclear how accurately current models recover plotted values from real scientific figures, as existing benchmarks rely largely on synthetic charts or cover only a limited range of chart types. 
We introduce \textbf{PlotGround}, an automated pipeline for building plot digitization benchmarks from real scientific figures and their author-released source data. 
PlotGround maps figures to source tables, identifies reconstructable panels, and generates quantitative questions with source-grounded reference values. 
We use PlotGround to construct \textbf{PlotGround-1k}, a human-verified benchmark of 1,119 questions from 1,066 bioRxiv preprints. Across sixteen multimodal models, the best reaches 87.5\% accuracy at a $\pm5\%$ relative-error tolerance. Tightening the tolerance to $\pm2\%$ lowers every model's accuracy by 11--24 percentage points, revealing a gap between approximate visual reading and precise quantitative recovery.
PlotGround's paired figure--source structure lets us compare how accurately the same values are recovered from figures and from source tables. Providing source tables instead of figures raises a coding agent's accuracy from 90.0\% to 97.4\% while cutting cost by 72\%.
\end{abstract}

\section{Introduction}

Scientific figures typically sit at the end of rich computational pipelines: raw measurements may be filtered, normalized, and statistically summarized before the resulting quantities are rendered visually \citep{peng2011reproducible, sandve2013ten, wilson2017good}. Yet when an AI agent reads a paper, pixels are often all it gets. Recovering the underlying numerical values from those pixels, known as plot digitization, is therefore a basic prerequisite for quantitative analysis over scientific figures \citep{savva2011revision, rohatgi2017webplotdigitizer, liu2023deplot}.

The reliability of this step becomes increasingly consequential as multimodal models and coding agents are deployed for scientific tasks. They are expected not only to ingest papers but also to verify reported findings and reproduce analyses \citep{skarlinski2024language, siegel2024core, starace2025paperbench}. Figures often contain key evidence supporting a paper's claims \citep{roberts2024scifibench, sun2026automated}, and even small digitization errors can silently propagate into subsequent analysis and conclusions.

Despite these stakes, we still know little about how accurately current models recover numerical values from real scientific figures. Existing digitization benchmarks offer limited guidance here because they have largely relied on synthetically rendered charts \citep{methani2020plotqa, hassan2023lineex, xia2025chartx, berkane2026self} or covered only a limited set of chart types \citep{luo2021chartocr, touhidul2026chartzero}. Synthetic rendering provides exact ground truth and controllable visual variation, but it does not fully capture the characteristics of real scientific figures, including multi-panel layouts, heterogeneous chart types, dense legends and annotations, and varied axis conventions. As a result, performance on these benchmarks may not reflect performance on figures encountered in scientific papers.

We introduce \textbf{PlotGround}, an automated pipeline for constructing plot digitization benchmarks directly from scientific papers and their author-released source data. 
Our key observation is that many papers already deposit the data behind their figures, but what is missing is an explicit mapping from source data to the plotted quantities. 
Unlike synthetic benchmarks that render numerical data into plots, PlotGround starts from real scientific figures and works backward. It reconstructs the plotted quantities from the corresponding source data to provide grounded reference values. 
The resulting figure--data pairs also allow us to query the same quantities from either the rendered figure or the source table, isolating how the representation of evidence affects agent performance (Section~\ref{sec:evidence}).

Building on PlotGround, we construct PlotGround-1k, a human-verified benchmark of 1,119 quantitative questions from 1,066 bioRxiv preprints. Our evaluation shows strong performance among frontier models: GPT-5.6 Sol reaches 87.5\% accuracy at a $\pm5\%$ tolerance, and four of the 16 models exceed 80\%. However, accuracy drops by 11--24 percentage points under a stricter $\pm2\%$ tolerance, indicating a remaining gap in numerical precision. During human verification, we also find 8 figures whose plotted values disagree with those reconstructed from their own deposited source data (Section~\ref{sec:discrepancies}). These cases suggest that accurate plot digitization could help flag inconsistencies between scientific figures and their source data.

Our contributions are threefold:
\begin{itemize}
    \item \textbf{A source-grounded construction pipeline.}
    We introduce PlotGround, an automated pipeline that turns real scientific figures into digitization benchmarks by grounding plotted values in author-released source data.

    \item \textbf{A realistic benchmark with paired evidence representations.} 
    We construct PlotGround-1k, a human-verified benchmark of 1,119 quantitative questions over real scientific figures. Each question is linked to the source table from which its reference value is computed, enabling controlled comparisons between figures and source data as evidence.

    \item \textbf{A characterization of the precision gap in scientific plot digitization.}
    We find that frontier models perform strongly under a $\pm5\%$ tolerance, but accuracy drops by 11--24 percentage points at $\pm2\%$. In an agentic setting, recovering the same quantities from source tables is more precise than estimating them from figures.

    % \item \textbf{Empirical findings and verification implications.} 
    % We demonstrate strong plot digitization performance among frontier models and document 8 figure--data discrepancies identified during human verification, highlighting the potential for model-assisted consistency checking.
\end{itemize}

\section{Related Work}

\paragraph{Scientific Agents.}
Recent work has increasingly explored AI agents for scientific tasks spanning literature synthesis \citep{asai2026synthesizing}, research-idea generation \citep{si2025can,baek2025researchagent}, experiment execution \citep{boiko2023autonomous,si2026towards}, and the replication or verification of published results \citep{hu2025repro,siegel2024core}. 
As these systems move beyond text-only interaction, they increasingly operate over heterogeneous scientific artifacts, including papers, code, tables, and figures \citep{khalighinejad2025matvix, lu2026towards}.
This creates a particular challenge for verification and reproducibility workflows, where recovering reported quantitative results may require locating and interpreting figures \citep{siegel2024core}.
Our work focuses on this interface, studying the extraction of numerical information from scientific figures and how its representation affects downstream agent performance.

\paragraph{Plot Digitization.}
Plot digitization aims to recover the numerical values encoded by visual marks in a rendered chart,
ranging from interactive tools \citep{10.1145/3025453.3025957, rohatgi2017webplotdigitizer} to fully automated extraction methods \citep{liu2023deplot, liu2023matcha, yang2025efficient}. However, existing extraction benchmarks often rely on automatically rendered charts or cover only a limited set of chart types \citep{hassan2023lineex, xia2025chartx}. PlotQA uses real-world tabular data but automatically generates the chart images \citep{methani2020plotqa}; ExcelChart400K uses authentic charts collected from public Excel files but covers only bar, line, and pie charts \citep{luo2021chartocr}. 
Newer benchmarks have begun to move toward real-world figures, but important limitations remain. ChartZero constructs a 1,000-chart benchmark mined from heterogeneous PDFs but restricts evaluation to line charts \citep{touhidul2026chartzero}. SPUR and CharXiv evaluate models on real scientific figures, but through multiple-choice or descriptive and reasoning questions rather than open-ended recovery of plotted values \citep{wang2024charxiv, ding2026decoding}. 
Our benchmark addresses these gaps by evaluating open-ended numerical recovery across diverse real scientific figures, with ground-truth values derived from author-released source data.

\section{PlotGround}
\label{sec:plot}

\begin{figure}[htbp]
    \centering
    \includegraphics[width=1.00\textwidth]{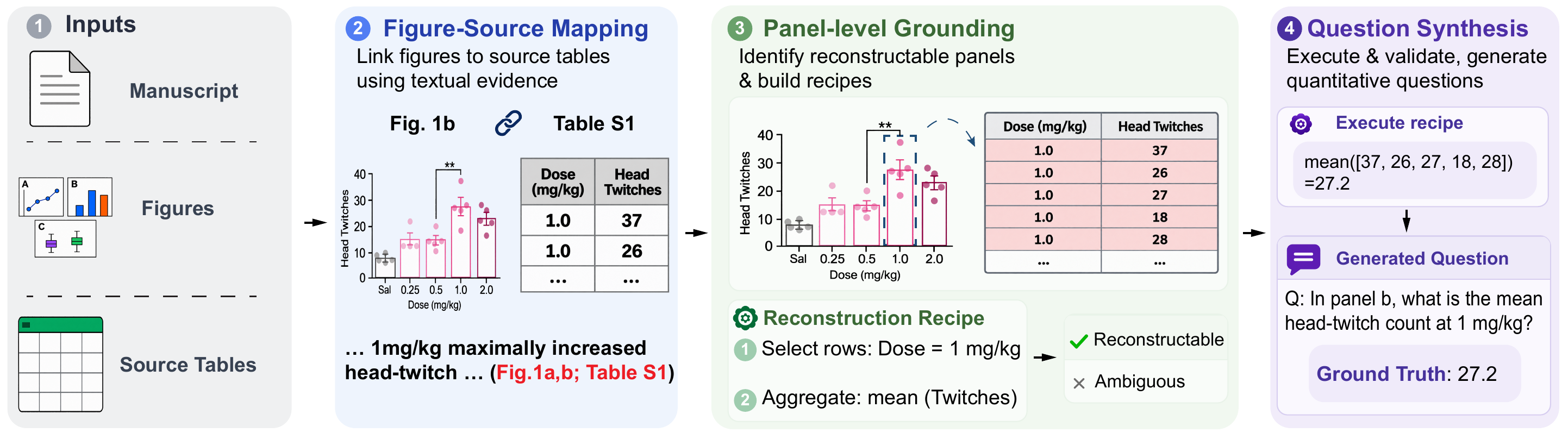}
\caption{
\textbf{Overview of the PlotGround pipeline.}
Given a bioRxiv paper organized as a filesystem (manuscript, figures, and supplementary source tables), \textbf{Mapping} proposes candidate figure--table bindings from textual evidence such as captions and in-text references; \textbf{Grounding} retains only panels whose values can be reliably reconstructed and records a reconstruction recipe, rejecting ambiguous ones; \textbf{Synthesis} follows the recipe, validates the recomputed values, and writes questions using only figure-visible labels. For clarity, the example shows a single panel with question text abbreviated. Real figures are typically multi-panel and are given to models in full (Appendix~\ref{app:bench_examples}).
}
    \label{fig: pipeline}
\end{figure}

\subsection{Pipeline Overview}

PlotGround turns a scientific figure and its corresponding source data into grounded quantitative questions. Each question refers to a visible element in the figure, such as a bar, labeled point, or box, and asks for its numerical value. The answer is recomputed from the underlying source table rather than read from the rendered image. We evaluate digitization at the level of \textbf{individual plotted values}, allowing each answer to be verified against its source data and scored independently.

PlotGround operates in three stages (Figure \ref{fig: pipeline}). \textbf{Mapping} identifies candidate source tables for each scientific figure. \textbf{Grounding} examines each mapped figure at the panel level, retaining only panels whose plotted values can be reliably reconstructed and recording a reconstruction recipe specifying the operations needed to recover them. \textbf{Synthesis} follows these recipes to compute reference values and generates panel-specific quantitative questions.

All three stages are executed by Claude Code \citep{anthropic_claude_code} using its default tool set, which allows the agent to navigate and read files, inspect spreadsheets, and execute code. We use Claude Opus 5 for the grounding stage and Claude Sonnet 5 for mapping and synthesis. Each paper is processed in an isolated working directory containing only the files associated with that paper.

\subsection{Figure--Source Mapping}
\label{sec:mapping}

The mapping stage identifies candidate supplementary source data for each scientific figure. The agent navigates the working directory and infers figure--source correspondences from the manuscript, captions, and spreadsheet summaries.

The agent proposes figure--source bindings supported by either \emph{direct} evidence, such as an explicit textual reference connecting a figure to a supplementary file, or \emph{indirect} evidence, such as a match between the figure description and the contents of a candidate table. When the available evidence explicitly identifies a specific panel, the correspondence may be resolved at the panel level; otherwise, it is retained at the figure level and further disambiguated during grounding.

\subsection{Panel-Level Grounding}
\label{sec:grounding}

Given a candidate figure--source mapping, the grounding stage determines which panels can be reliably reconstructed from the linked source data. Grounding is intentionally conservative: a panel is retained only when its plotted quantity can be recovered with simple, well-defined operations and when the correspondence between visual elements and source-table entries is unambiguous.

The stage first applies a \emph{reliability gate}. We retain quantitative charts whose plotted values can be recovered from the source table using basic operations such as selection, aggregation, ratios, and logarithmic transformations. Panels are rejected when reconstruction requires domain-specific analyses, lacks figure-visible identifiers for question construction, or leaks exact values through printed annotations.

For each retained panel, the agent maps visual groups to explicit rows or columns in the source table and records a structured reconstruction recipe specifying the relevant sheet, columns, row filters, and any aggregation or normalization. Every mapping must be anchored to literal evidence, such as a table header, a category-column value, or a token visible in the figure. Mappings based on domain priors, conventional ordering, or sample-ID ordinals are disallowed. Panels with ambiguous column choices (e.g., raw versus adjusted \textit{p}-values), unidentified normalization references, or incomplete source-data coverage are also rejected.

\subsection{Question Synthesis}
\label{sec:synthesis}

The synthesis stage follows the recorded reconstruction recipe on the source table to recompute the plotted values. It then matches these values to the individually readable elements of each panel: bar heights, line or survival values at labeled x positions, the marked statistics of a box plot (median and quartiles), and the coordinates of in-figure-labeled points in scatter and volcano plots. 

Before generating questions, it checks that the recipe's assumptions are supported by explicit evidence in the source file or paper. It also verifies that the recomputed values satisfy basic constraints of the plotted quantity (e.g., survival probabilities must be non-increasing over time). As a final guard against incorrect figure--source mappings, the agent reads each target off the figure and discards it only when the reading differs from the recomputed value by a large margin (at least three times the $\pm5\%$ tolerance used for evaluation).
Among the elements that pass these checks, it selects the clearest target in each panel, favoring large, uncrowded marks.

Each question asks for a single numerical value and names its target using only identifiers visible in the figure, such as the panel label, an x-tick or legend entry, and the axis label and units, e.g., ``In panel b, what is the mean head-twitch count at 1 mg/kg?'' (Figure~\ref{fig: pipeline}). This ensures that questions can be answered from the figure alone without access to the source data. To reduce redundancy, synthesis emits at most two questions per figure, drawn from different panels. Additional examples, shown with their full multi-panel figures and spanning different chart types, are provided in Appendix~\ref{app:bench_examples}.

\section{Experiments}

\subsection{Benchmark Construction}
\label{sec:corpus}

\paragraph{Paper selection.} We build PlotGround on the 2025 bioRxiv corpus available through Paperclip\footnote{\url{https://paperclip.gxl.ai/}}, an agent-native interface to scientific literature that exposes papers in a filesystem-like environment. This representation gives coding agents direct access to the components of a paper, including the manuscript, figures, and deposited supplementary files, and allows them to navigate and reason over these artifacts programmatically. We focus on bioRxiv because source-data deposition is common in biology, making it possible to link published figures to the underlying tables from which their values can be reconstructed.

We require each paper to contain at least one machine-readable tabular supplement (\texttt{.xlsx}, \texttt{.xls}, \texttt{.csv}, or \texttt{.tsv}), as these formats provide structured source data that is well suited to direct figure reconstruction. Among 37,635 preprints with non-empty supplements, 10,540 satisfy this criterion, and all of them enter the pipeline.
\paragraph{Pipeline funnel.}
The corpus then passes through the three stages in Sections \ref{sec:mapping}--\ref{sec:synthesis}. 
\emph{Mapping} identifies 28,536 figure--source bindings across 8,197 papers. To control cost, we keep at most one mapping per paper and randomly sample 7,487 of the 8,197 papers for downstream processing.
\emph{Grounding} evaluates the sampled mappings, covering 22,541 figure panels in total. Among them, 3,894 panels from 1,568 figures admit a valid reconstruction recipe.
\emph{Synthesis} generates at most two questions per figure, and figures that do not admit a suitable question yield none. When a figure has two questions targeting panels of the same chart type, we remove one at random, leaving 1,205 candidates. We manually audit all of them and retain 1,119 verified questions from 1,066 papers, which constitute \textbf{\ourbench{}} (see Appendix~\ref{app:human_eval_protocol} for the verification protocol). Table~\ref{tab:funnel} summarizes the resulting counts.

\begin{table}[htb!]
\small
\caption{PlotGround pipeline funnel. Counts are given in the unit of each stage (papers, mappings, panels, or questions), since one paper can yield multiple figure–source mappings and one figure multiple panels.}
\label{tab:funnel}
\begin{center}
\begin{tabular}{llr}
\toprule
Stage & Step & Retained \\
\midrule
Collection & Index 2025 preprints with supplements & 37,635 papers \\
Filtering & Require a tabular supplement & 10,540 papers \\
Mapping    & Bind figures to source tables (Section \ref{sec:mapping}) & 28,536 mappings (8,197 papers) \\
Downsampling & Randomly sample papers, one mapping per paper & 7,487 mappings \\
Grounding  & Admit reconstructable panels (Section \ref{sec:grounding}) & 3,894 panels (1,568 figures) \\
Synthesis  & Generate and filter questions (Section \ref{sec:synthesis}) & 1,205 questions\\
Verification & Human audit of all candidates & \textbf{1,119 questions} (1,066 papers) \\
\bottomrule
\end{tabular}
\end{center}
\end{table}

\subsection{Experimental Setup}

We evaluate sixteen frontier multimodal models on \ourbench{} under a shared protocol. Given the full figure image and a quantitative question targeting a specific panel (see Appendix \ref{app:prompt-eval} for the full prompt), each model predicts the corresponding numerical value from the figure alone. 
We convert all figures to PNG, cap the longer edge at 1,600 pixels, and preserve the original aspect ratio. Following the relaxed accuracy criterion used in PlotQA \citep{methani2020plotqa}
and ChartQA \citep{masry2022chartqa}, we measure accuracy at a relative-error tolerance of $\pm5\%$ and additionally at a stricter $\pm2\%$ tolerance to assess numerical precision. We also report the average cost and output-token count per question.

\textbf{Models.} We evaluate sixteen models spanning nine developers:
OpenAI's GPT-5.6 Sol, GPT-5.6 Luna, and GPT-5.6 Terra
\citep{openai2026gpt56};
Google's Gemini 3.8 Flash, Gemini 3.7 Flash, and Gemini 3.1 Pro
\citep{google2026gemini31pro,google2026gemini37flash,google2026gemini38flash};
Anthropic's Claude Opus 5 and Claude Sonnet 5
\citep{anthropic2026opus5,anthropic2026sonnet5};
Alibaba's Qwen3.8 Max and Qwen3.8 Flash \citep{qwen2026qwen38max,qwen2026qwen38flash};
Moonshot's Kimi K3 \citep{kimi2026k3};
Z.ai's GLM-5.3 Flash \citep{zai2026glm53flash};
MiniMax's M3 \citep{minimax2026m3};
DeepSeek's V4.1 Flash and V4 Flash Vision Exp
\citep{deepseek2026v41flash,deepseek2026v4flashvision};
and Thinking Machines Lab's Inkling
\citep{thinkingmachines2026inkling}.
Each model is evaluated independently through OpenRouter\footnote{\url{https://openrouter.ai/}} with its default settings. We do not override sampling parameters, output limits, or reasoning configuration. All models returned a numeric answer for every question.

\subsection{Main Results}
\label{sec:results}
\begin{table}[htb!]
\caption{Model accuracy on PlotGround-1k (1,119 questions). A prediction counts as correct if its relative error with respect to the gold value is within the tolerance. All models were run with their default settings, including reasoning configuration. Cost and output tokens are averaged over all questions; output tokens include reasoning tokens where the provider reports them.
}
\label{tab:results}
\begin{center}
\begin{tabular}{@{}lrrrr@{}}
\toprule
Model & Acc.\ ($\pm$2\%) & Acc.\ ($\pm$5\%) & Avg.\ Cost (\textcent) & Avg.\ Output tokens \\
\midrule
GPT-5.6 Sol & \textbf{73.8} & \textbf{87.5} & 1.013 & 419 \\
GPT-5.6 Luna & 69.7 & 83.8 & 0.113 & 447 \\
GPT-5.6 Terra & 66.6 & 82.1 & 0.868 & 231 \\
Gemini 3.7 Flash & 65.7 & 81.2 & 0.428 & 909 \\
Gemini 3.8 Flash & 65.5 & 79.8 & 0.933 & 2,257 \\
Qwen3.8 Max & 57.7 & 77.9 & 2.945 & 4,269 \\
Claude Opus 5 & 60.2 & 77.1 & 2.724 & 564 \\
Claude Sonnet 5 & 55.6 & 74.4 & 0.631 & 105 \\
Gemini 3.1 Pro & 49.9 & 67.6 & 4.237 & 3,337 \\
Qwen3.8 Flash & 42.6 & 64.4 & 0.223 & 4,111 \\
Kimi K3 & 38.2 & 62.5 & 8.371 & 5,545 \\
GLM-5.3 Flash & 39.8 & 60.2 & 0.105 & 1,495 \\
Inkling & 32.0 & 50.3 & 0.866 & 1,633 \\
MiniMax M3 & 33.8 & 50.3 & 0.102 & 358 \\
DeepSeek V4.1 Flash & 29.3 & 47.0 & 0.131 & 904 \\
DeepSeek V4 FV exp & 17.0 & 28.6 & 0.261 & 3,177 \\
\bottomrule
\end{tabular}  
\end{center}
\end{table}

Table~\ref{tab:results} shows that several frontier models perform strongly on \ourbench{}: at a $\pm5\%$ relative-error tolerance, four models exceed 80\% accuracy, and GPT-5.6 Sol leads at 87.5\%. At the stricter $\pm2\%$ tolerance, accuracy drops by 11--24 percentage points across models, with GPT-5.6 Sol falling to 73.8\%. The overall ranking remains broadly stable, suggesting that relative model performance is robust to this change in tolerance. The drop still matters when extracted values are reused in downstream calculations: errors can affect derived quantities such as ratios
or fitted parameters, so precision sufficient for approximate
inspection may not suffice for quantitative reuse. A breakdown of accuracy by chart type is provided in the Appendix ~\ref{app:accuracy_chart_type}.

\begin{figure}[htb!]
\centering
\includegraphics[width=\linewidth]{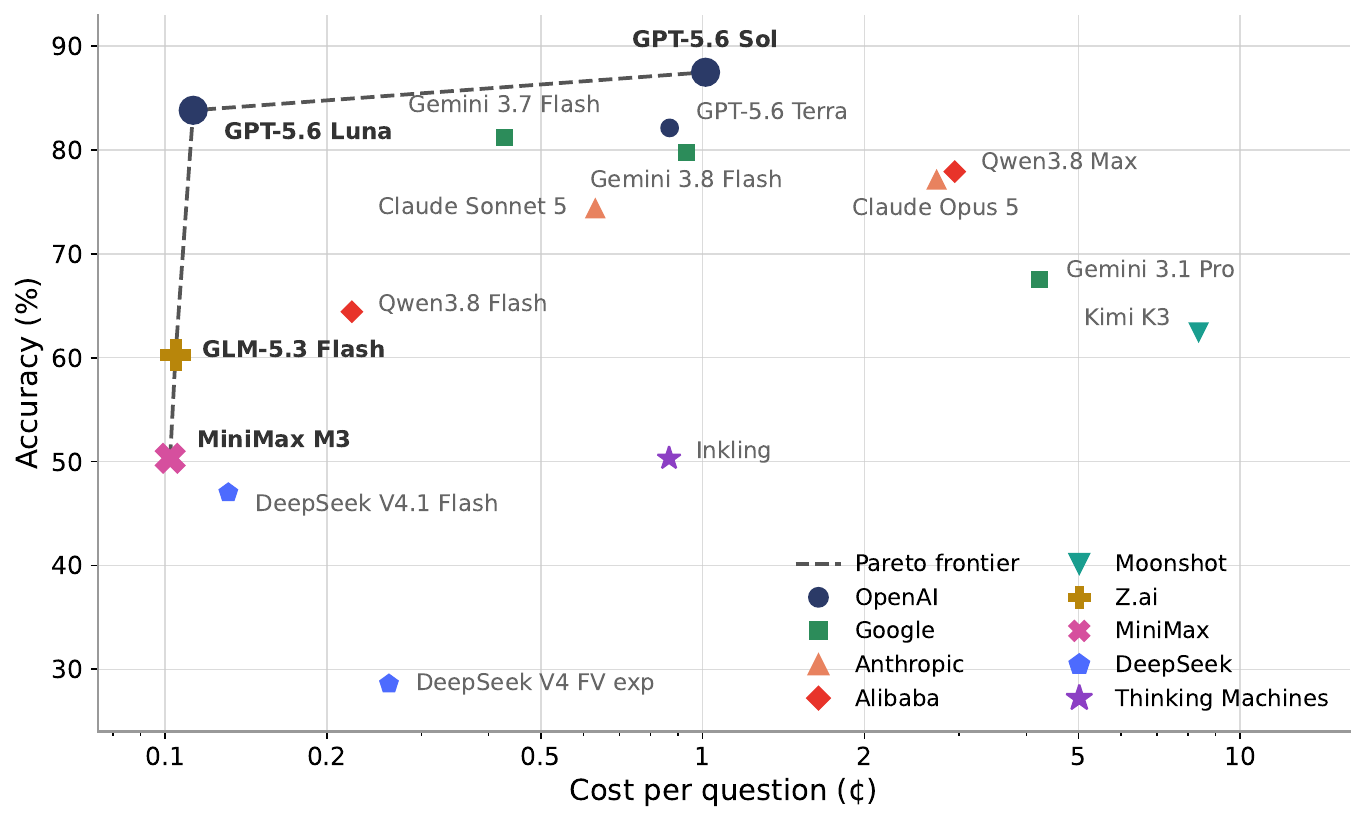}
\caption{
\textbf{Accuracy versus inference cost on PlotGround-1k.}
Each point represents a model, with colors and marker shapes indicating developers.
Accuracy is measured at a $\pm5\%$ tolerance, and mean inference cost per question is shown in cents on a logarithmic scale.
The dashed line connects Pareto-optimal models, whose labels are shown in bold.
Higher cost does not necessarily yield higher accuracy: GPT-5.6 Luna achieves 83.8\% accuracy at 0.11 cents per question, compared with 62.5\% at 8.37 cents for Kimi K3.
}
\label{fig:cost_accuracy}
\end{figure}

Higher accuracy does not require higher inference cost
(Figure~\ref{fig:cost_accuracy}). GPT-5.6 Luna achieves 83.8\%
accuracy at 0.113\textcent{} per question, approximately 4\%
of the cost of Claude Opus 5 while scoring 6.7 percentage points
higher. The two most expensive models, Kimi K3 and Gemini 3.1 Pro,
both rank in the bottom half despite averaging 5,545 and
3,337 reported output tokens per question, respectively.

\subsection{Qualitative Error Analysis}

To better understand where current models fail, we manually inspect representative incorrect predictions across the evaluated models. We observe several recurring error patterns that are not apparent from aggregate accuracy alone. In particular, many failures arise from converting visual position into precise numerical values, especially when interpolation between labeled ticks is required.

\begin{figure}[t]
\centering
\includegraphics[width=\linewidth]{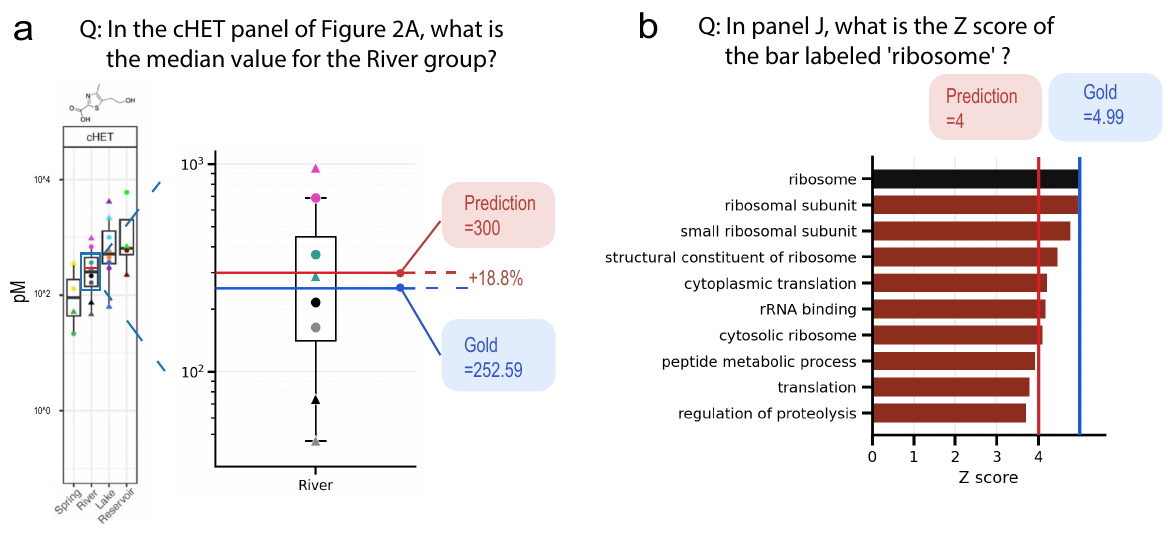}
\caption{\textbf{Representative qualitative failure cases.} (a) On a log axis, a visually small discrepancy in position can correspond to a substantial numerical error; the model predicts 300 for a gold value of 252.59. (b) Models often snap predictions to labeled tick marks, failing to account for the target’s position between or beyond ticks. For a bar with gold value 4.99, eight of sixteen models predict the final labeled tick at 4.}
\label{fig:failure_cases}
\end{figure}

\paragraph{Logarithmic axes.} Reading a value from a logarithmic axis requires geometric interpolation between labeled ticks, which is a frequent source of error among the cases we inspected. Models generally recognize that the axis is logarithmic, but struggle to convert a target's visual position into the correct numerical value when it falls between labeled ticks. On a log scale, small positional errors can translate into much larger relative numerical errors because equal distances correspond to equal ratios between values. For example, predicting $300$ for a gold value of $252.59$ yields a relative error of approximately $19\%$ (Figure~\ref{fig:failure_cases}a), despite the prediction appearing visually close to the plotted point.

\paragraph{Snapping to tick marks.}
Predictions often concentrate on labeled tick values, even when the target lies visibly between or beyond them. In a horizontal bar chart (Figure~\ref{fig:failure_cases}b) whose $x$-axis is labeled at $0$, $1$, $2$, $3$ and $4$, the target bar has a gold value of $4.99$ and visibly extends past the final gridline. Eight of sixteen models nevertheless answered exactly $4$, the last labeled tick. Only three read the overshoot and answered $5$, two gave intermediate values ($4.2$ and $4.5$), and the remaining three were further off ($3$, $20$, and $40$). 
This behavior is consistent with models estimating the target relative to the nearest tick and then reporting that tick, or a nearby round number. As a result, accuracy may depend on where the gold value lies relative to the axis labels: values close to ticks are often recovered correctly, whereas values between ticks are more error-prone.

\subsection{Effect of Evidence Representation}
\label{sec:evidence}

The evaluations above measure plot digitization in its purest form: a model is shown a figure and a question, without access to the surrounding paper or source data. Recent work on scientific agents, however, has begun to move beyond the paper as a single static document, exposing agents to broader collections of research artifacts \citep{miao2025paper2agent, liu2026last}. In such settings, the same quantitative evidence may also be available in structured source data. This raises a basic question for agentic scientific reasoning: how much does the representation of quantitative evidence affect an agent's ability to recover and use it?

PlotGround provides a natural way to study this question because every benchmark instance links a plotted quantity to the source table from which its value is reconstructed. We can therefore hold the scientific question fixed while varying whether the agent has access to figures, source data, or both. 
In particular, we ask what the agent gains when the same quantity can be recovered from structured data rather than estimated from pixels.

\paragraph{Setup.} We rewrite questions from PlotGround-1k in the paper's own terms. References to figures, panels, axes, and tables are removed so that the same question can be answered using either the figure or the underlying source data. For example, ``In panel A, what is the log2 fold change of the point labeled CHAC1?'' becomes ``In the RNAseq differential expression analysis (DESeq2, compared to vehicle-treated cells) of HEK293T cells treated for 6 h with MBX-2982 (10 $\mu$M), what is the log2 fold change in expression of CHAC1?'' (see Appendix~\ref{app:prompt-rewrite} for the rewriting prompt). During rewriting, we also flag and discard questions whose answers are stated directly in the manuscript text. The evaluation set consists of 500 randomly sampled questions that pass this filter. 

Claude Code (Claude Sonnet 5) then answers each question under three evidence conditions. Each run takes place in a Docker container holding the paper's files, with the same prompt throughout (Appendix \ref{app:evidence-eval}). Only the contents of the working directory differ between conditions. 
All three conditions include the full text (the article XML from the bioRxiv
submission package). The agent additionally receives figure images
(\textbf{Figure}), supplementary data files (\textbf{Table}), or both
(\textbf{Both}). The manuscript PDF is withheld throughout because it contains the rendered figures. This setting differs from the single-shot protocol of Table~\ref{tab:results} in question wording, available context, and image resolution, so the two accuracies are not directly comparable.

\begin{table}[t]
\caption{\textbf{Effect of evidence representation on agent performance.}
Claude Code (Claude Sonnet 5) answers the same 500 rewritten questions under
three evidence conditions: figures, source tables, or both. Accuracy is
measured at a $\pm5\%$ relative-error tolerance; cost and latency are averaged
over all questions.}
\label{tab:evidence}
\begin{center}
\begin{tabular}{@{}lrrr@{}}
\toprule
Condition & Accuracy ($\pm$5\%) & Avg.\ Cost (\$) & Avg.\ Latency (s) \\
\midrule
\textbf{Figure} (full text + figures)                  & 90.0          & 0.408 & 139 \\
\textbf{Table}  (full text + source tables)            & 97.4          & 0.113 &  51 \\
\textbf{Both}   (full text + figures + source tables)  & \textbf{98.2} & 0.115 &  51 \\
\bottomrule
\end{tabular}
\end{center}
\end{table}

\paragraph{Source data is more accurate, more precise, and cheaper.}
Access to structured source data substantially improves both accuracy and
efficiency (Table~\ref{tab:evidence}). With the figure alone, Claude Code reaches 90.0\% accuracy, but each question costs \$0.41 and takes 139\,s on average. The agent takes about 26 turns per question, repeatedly cropping and enlarging the target panel and, in most runs, measuring pixel positions directly from the image array. Replacing figures with source tables raises accuracy to 97.4\%, while reducing cost by 72\% and latency by 63\%. Among answers that pass at $\pm$5\%, the median relative error is 0.39\% from the figure against 0.001\% from the table.

\paragraph{Given both, the agent predominantly relies on the table.}
Adding figures alongside the tables provides little additional benefit: accuracy increases only marginally to 98.2\%, while cost and latency remain nearly identical to the table-only condition. The agent's working traces in the \textbf{Both} condition help explain this result. It accesses the source table for 492 of the 500 questions, but inspects a figure image for only 20. Among these 20 cases, 13 also involve table access, usually before the figure is inspected, suggesting that the image is used primarily to confirm a value already obtained from the table. Thus, when both representations are available, the agent overwhelmingly treats structured data as the primary source of quantitative evidence.

\subsection{Figure--Data Discrepancies}
\label{sec:discrepancies}

\begin{figure}[htbp]
  \centering
  \includegraphics[width=\textwidth]{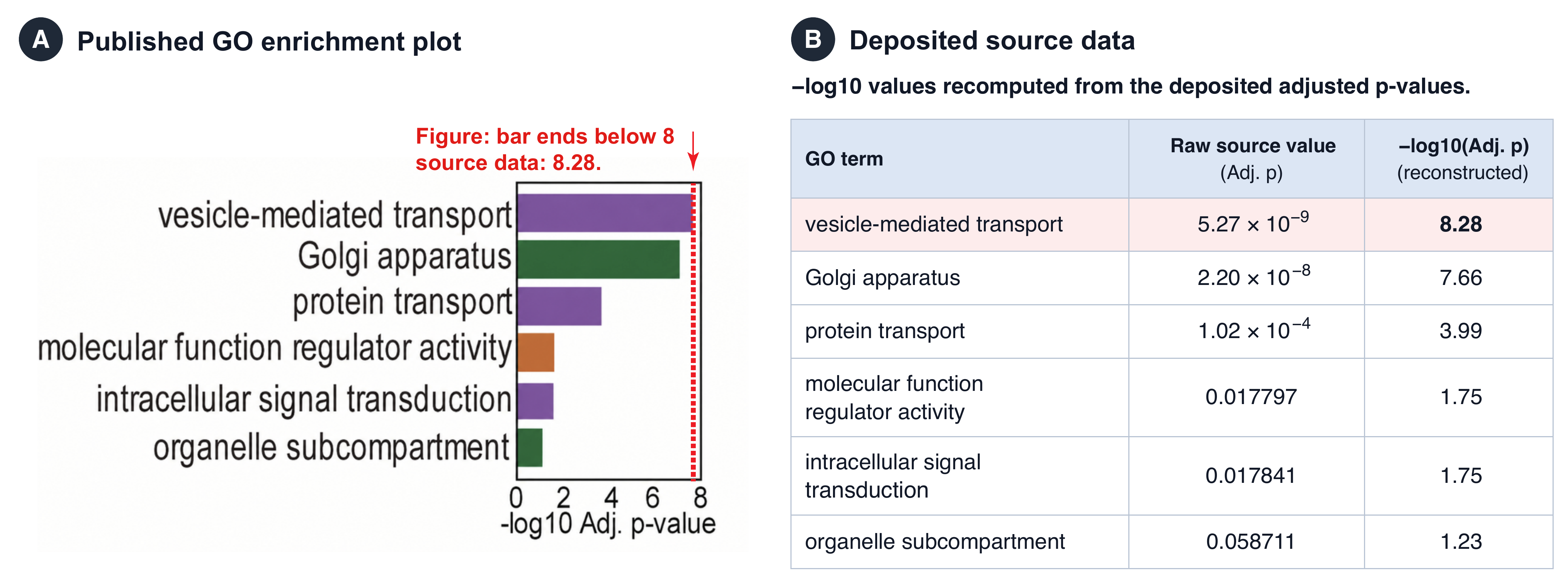}
  \caption{
  \textbf{Example of a discrepancy between a published figure and its deposited source data.}
  The published GO enrichment plot differs from the values
  reconstructed from the deposited source data. The discrepancy is clearest for the most significant term. The deposited adjusted $p$-value for vesicle-mediated transport ($5.27 \times 10^{-9}$) corresponds to $-\log_{10} = 8.28$, but the bar shown for this term ends at approximately 7.7, short of the axis limit at 8. Axis clipping cannot explain the gap, since a clipped bar would extend to the limit.}
  \label{fig:go}
\end{figure}

During verification, we found 8 candidate questions (from 8 figures) with apparent discrepancies between the plotted values and those reconstructed from the associated source data. Our review did not identify errors in the reconstruction recipes that could account for these differences. Examples included a bar ending below 8 despite a reconstructed value of 8.28 (Figure \ref{fig:go}), and a survival curve whose genotype labels appeared inconsistent with the source data (Supp Figure \ref{fig:survival}). We could not reconcile these differences using the available source data and manuscript descriptions, so we excluded these candidates from \ourbench{}.

Although identifying figure–data discrepancies was not an objective of the pipeline, these cases suggest a practical application. Given that frontier models can often recover plotted values accurately (Section~\ref{sec:results}), the same three steps (mapping a figure to its source table, reconstructing the plotted quantity, and comparing it with the value read from the figure) could be used to flag potential figure--data inconsistencies for further inspection. On the 8 discrepant figures, GPT-5.6 Sol's reading agrees with the figure rather than the table in every case. Since a model's own reading errors also trigger such flags, the check currently serves best as a filter that narrows down candidates for human review, and it becomes more precise as reading accuracy improves.
\section{Conclusion}
\label{conclusion}

We introduced PlotGround, a pipeline for constructing plot digitization benchmarks from real scientific figures and their source data. The resulting benchmark, PlotGround-1k, contains 1,119 human-verified questions. Evaluation of 16 multimodal models reveals a remaining gap between approximate visual reading and precise numerical recovery. In our agentic evaluation, replacing figures with source tables improves accuracy while substantially reducing cost, highlighting the value of structured data for quantitative scientific workflows. We also identify figures whose plotted values disagree with their own deposited data, suggesting figure--data consistency checking as a downstream use of accurate plot digitization.

\paragraph{Limitations.} PlotGround-1k is drawn from bioRxiv preprints that deposit tabular source data, so its current coverage is concentrated in biology and may not reflect figure conventions in other fields. However, the pipeline itself is not tied to bioRxiv. Journals that publish per-figure source data, such as those in the Nature Portfolio, could support similar benchmark construction in other disciplines. Our conservative grounding criteria and preference for clearly readable targets help ensure the reliability of the pipeline, but favor figures whose plotted values are relatively easy to verify. Performance on PlotGround-1k may therefore not generalize to the full range of scientific figures. In addition, although the pipeline is automated, PlotGround-1k is fully human-verified (the audit accepted 92.9\% of candidates), and reducing the need for such review remains open. Finally, our evaluation focuses on recovering individual plotted values, an atomic operation in full plot digitization. PlotGround's explicit links between figures and source data could facilitate extending this evaluation to complete data series. We leave these extensions to future work.
\bibliography{iclr2027_conference}
\bibliographystyle{iclr2027_conference}

\clearpage
\newpage

\appendix

\section{Benchmark Details}

\subsection{Benchmark Composition}
\begin{table}[htpb!]
\caption{\textbf{Chart-type distribution of \ourbench{}.}
The benchmark contains 1,119 questions across 10 chart types.
Bar, volcano, and scatter plots together account for 77.3\% of the questions.
The remaining types include line plots, distribution plots, and survival curves.}
\label{tab:charttypes}
\begin{center}
\begin{tabular}{@{}lrr@{}}
\toprule
Chart type & Questions & Share (\%) \\
\midrule
Bar          & 423 & 37.8 \\
Volcano      & 268 & 23.9 \\
Scatter      & 174 & 15.5 \\
Box          & 112 & 10.0 \\
Line         &  86 &  7.7 \\
Violin       &  27 &  2.4 \\
Bubble       &  15 &  1.3 \\
MA plot      &   6 &  0.5 \\
Strip        &   5 &  0.4 \\
Survival     &   3 &  0.3 \\
\midrule
Total        & 1{,}119 & 100.0 \\
\bottomrule
\end{tabular}
\end{center}
\end{table}

\subsection{Benchmark Examples}
\label{app:bench_examples}

\begin{examplebox}{Box Plot}
  {\centering
    \includegraphics[width=\linewidth]{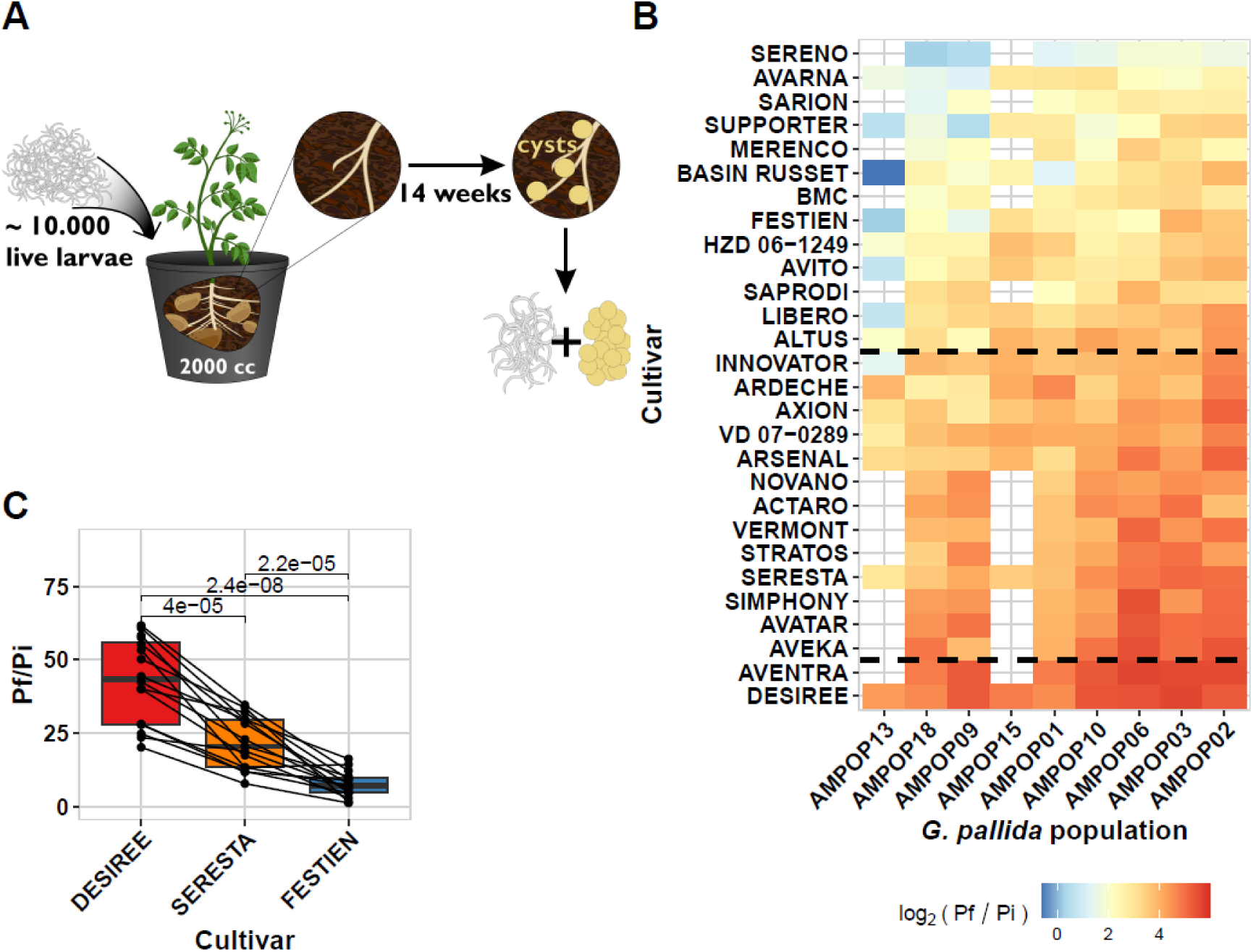}
    \par
  }
  \par\smallskip
  \textbf{Question:} In panel C, what is the median (center line) of the DESIREE box? (y-axis: 'Pf/Pi') \\
  \textbf{Answer:} 43.147
\end{examplebox}

\begin{examplebox}{Bar Plot}
  {\centering
    \includegraphics[width=\linewidth]{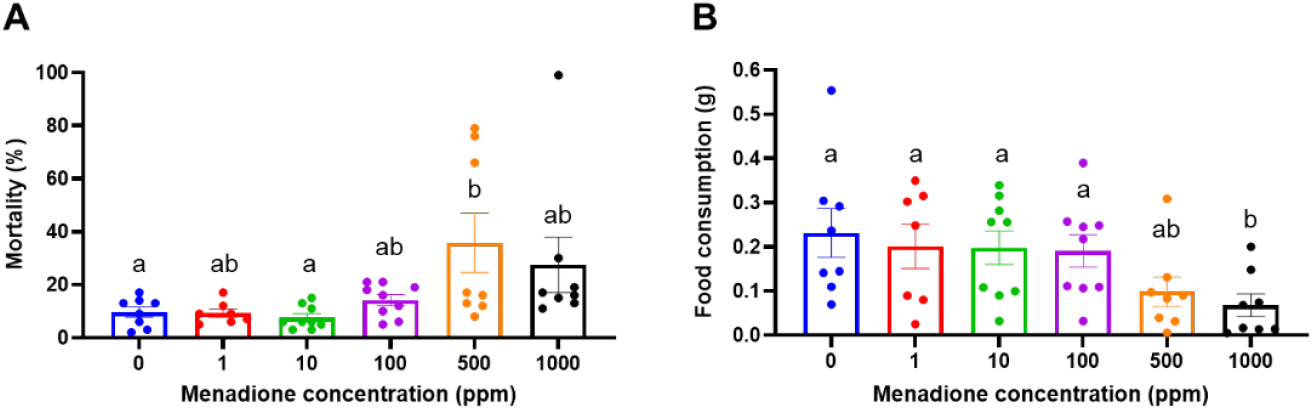}
    \par
  }
  \par\smallskip
  \textbf{Question:} In panel B, what is the bar height for the Menadione concentration of 500 ppm? (y-axis: 'Food consumption (g)')\\
  \textbf{Answer:} 0.098525
\end{examplebox}

\begin{examplebox}{Scatter Plot}
  {\centering
    \includegraphics[width=\linewidth]{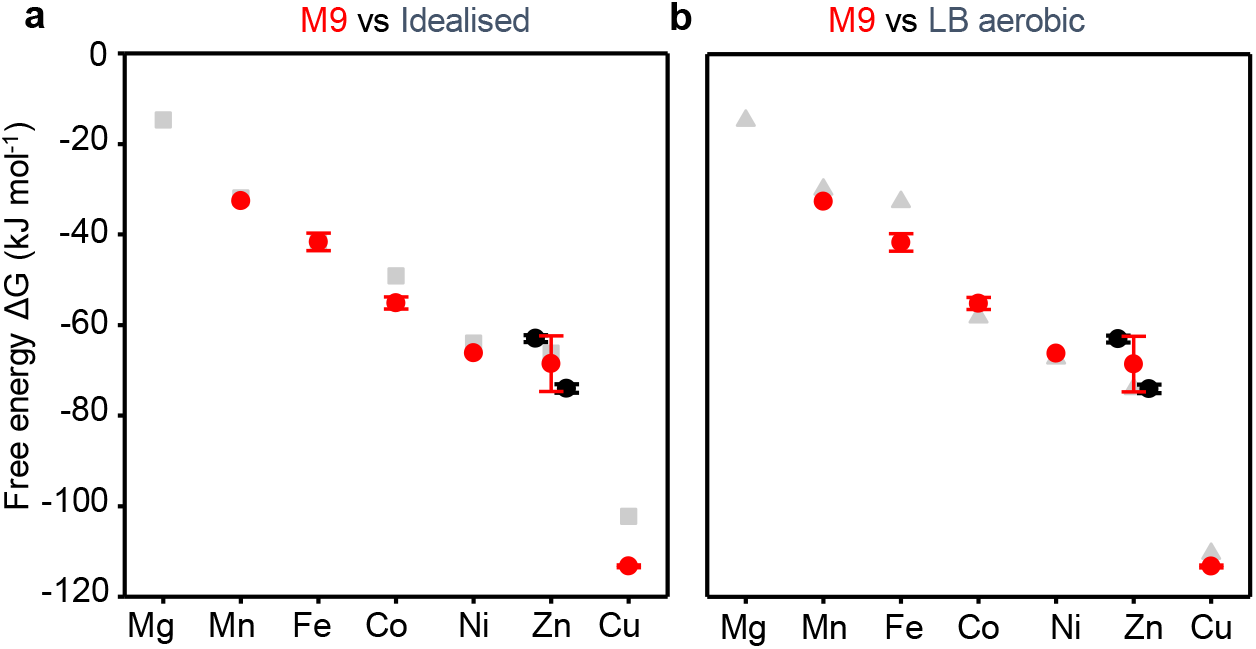}
    \par
  }
  \par\smallskip
  \textbf{Question:} In panel a, what is the free energy $\Delta G$ (kJ mol$^{-1}$) of the Idealised (gray square) data point at the Mg position?\\
  \textbf{Answer:} -14.698
\end{examplebox}

\begin{examplebox}{Line Plot}
  {\centering
    \includegraphics[width=\linewidth]{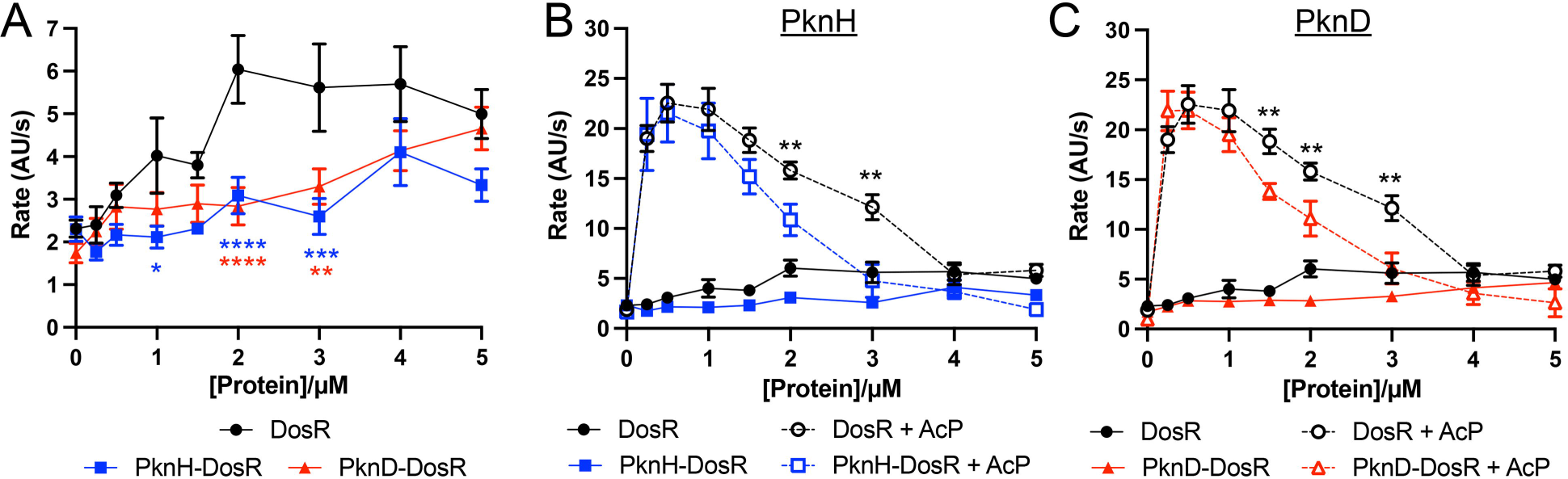}
    \par
  }
  \par\smallskip
  \textbf{Question:} In panel A, what is the Rate (AU/s) of the DosR series at [Protein]/µM = 2?\\
  \textbf{Answer:} 6.04
\end{examplebox}

\begin{examplebox}{Volcano Plot}
  {\centering
    \includegraphics[width=\linewidth]{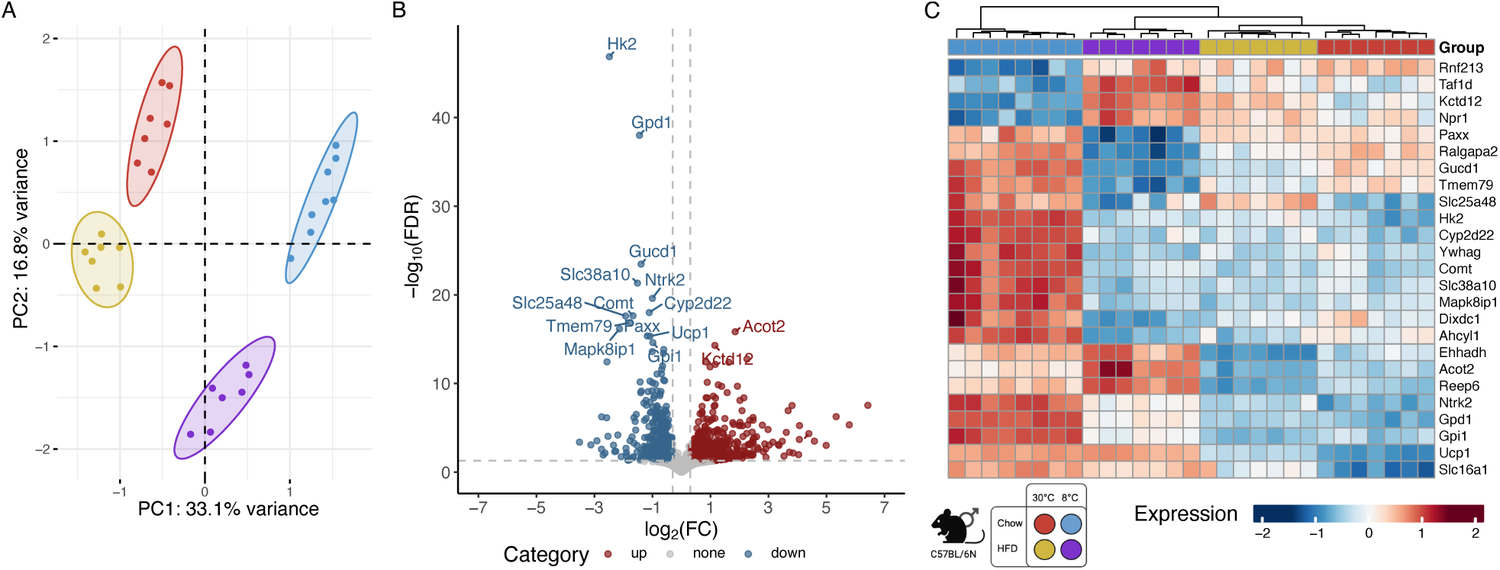}
    \par
  }
  \par\smallskip
  \textbf{Question:} In panel B, what is the -log10(FDR) value (y-axis) of the point labeled "Hk2" in the volcano plot?\\
  \textbf{Answer:} 46.86007339557034
\end{examplebox}

\begin{examplebox}{Strip Plot}
  {\centering
    \includegraphics[width=\linewidth]{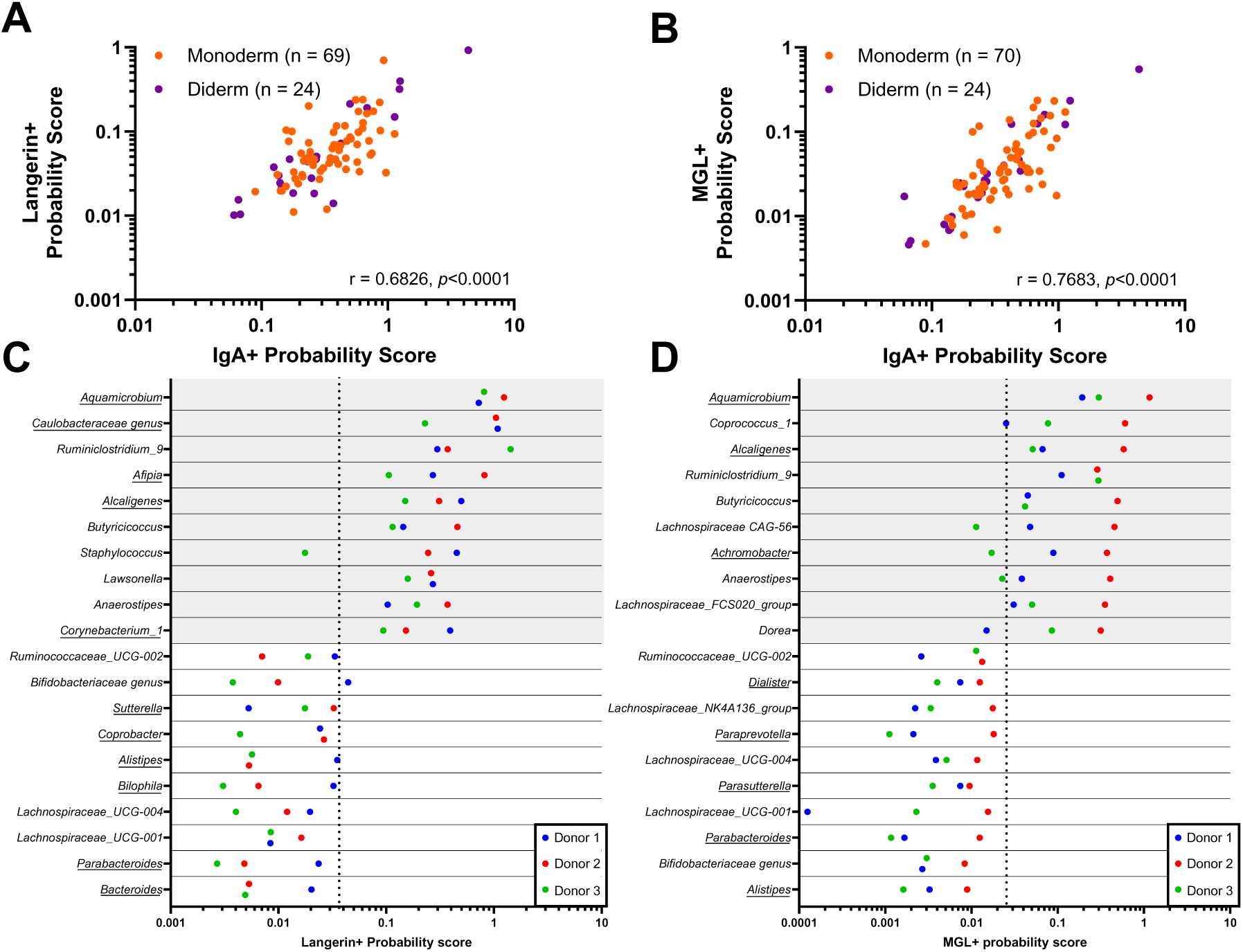}
    \par
  }
  \par\smallskip
  \textbf{Question:} In panel D, what is the MGL+ probability score for the Donor 2 (red) dot in the Aquamicrobium row?\\
  \textbf{Answer:} 1.165311
\end{examplebox}

\begin{examplebox}{Bubble Plot}
  {\centering
    \includegraphics[width=\linewidth]{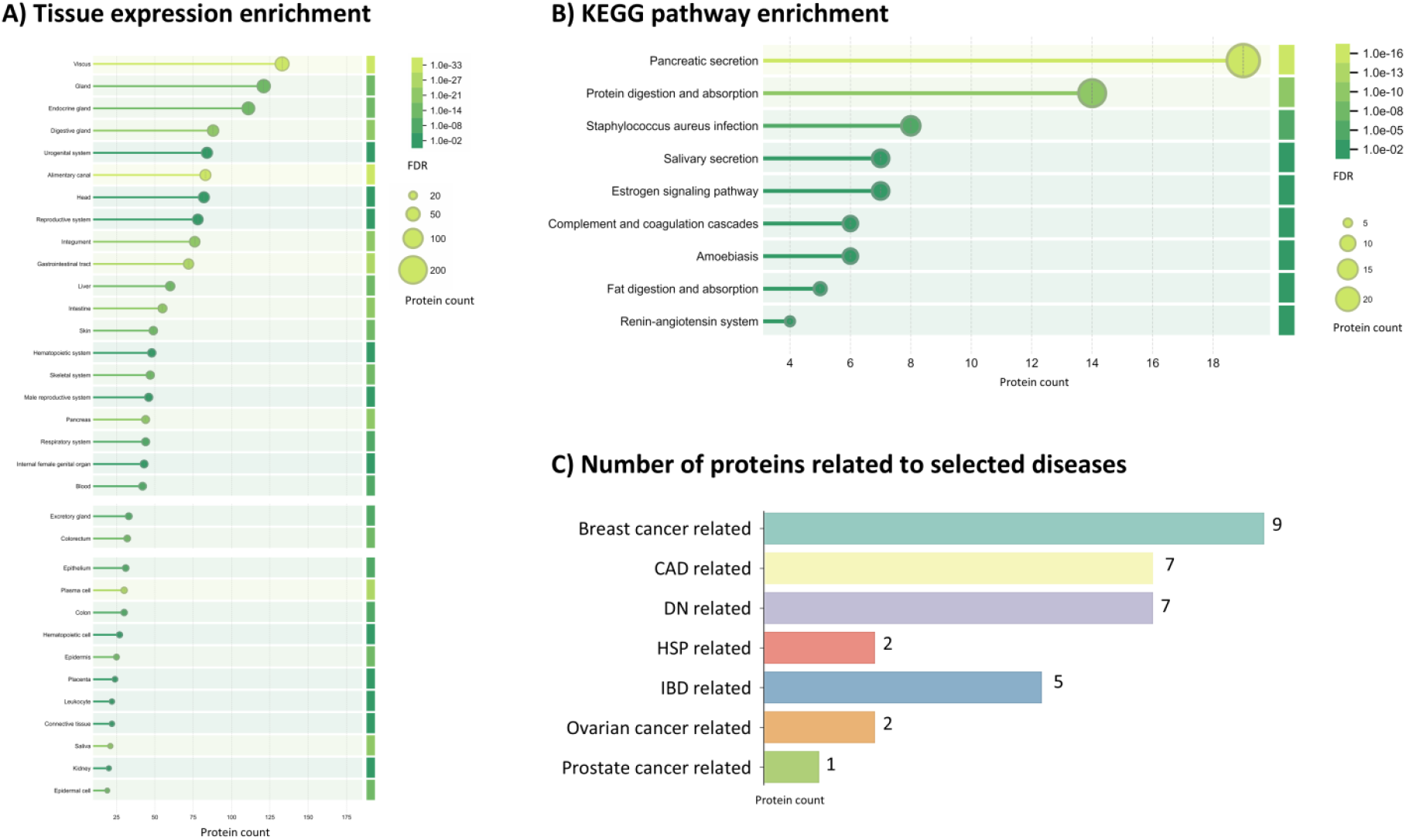}
    \par
  }
  \par\smallskip
  \textbf{Question:} In panel A, what is the protein count for the bubble labeled 'Viscus'?\\
  \textbf{Answer:} 132
\end{examplebox}

\begin{examplebox}{Survival Curves}
  {\centering
    \includegraphics[width=\linewidth]{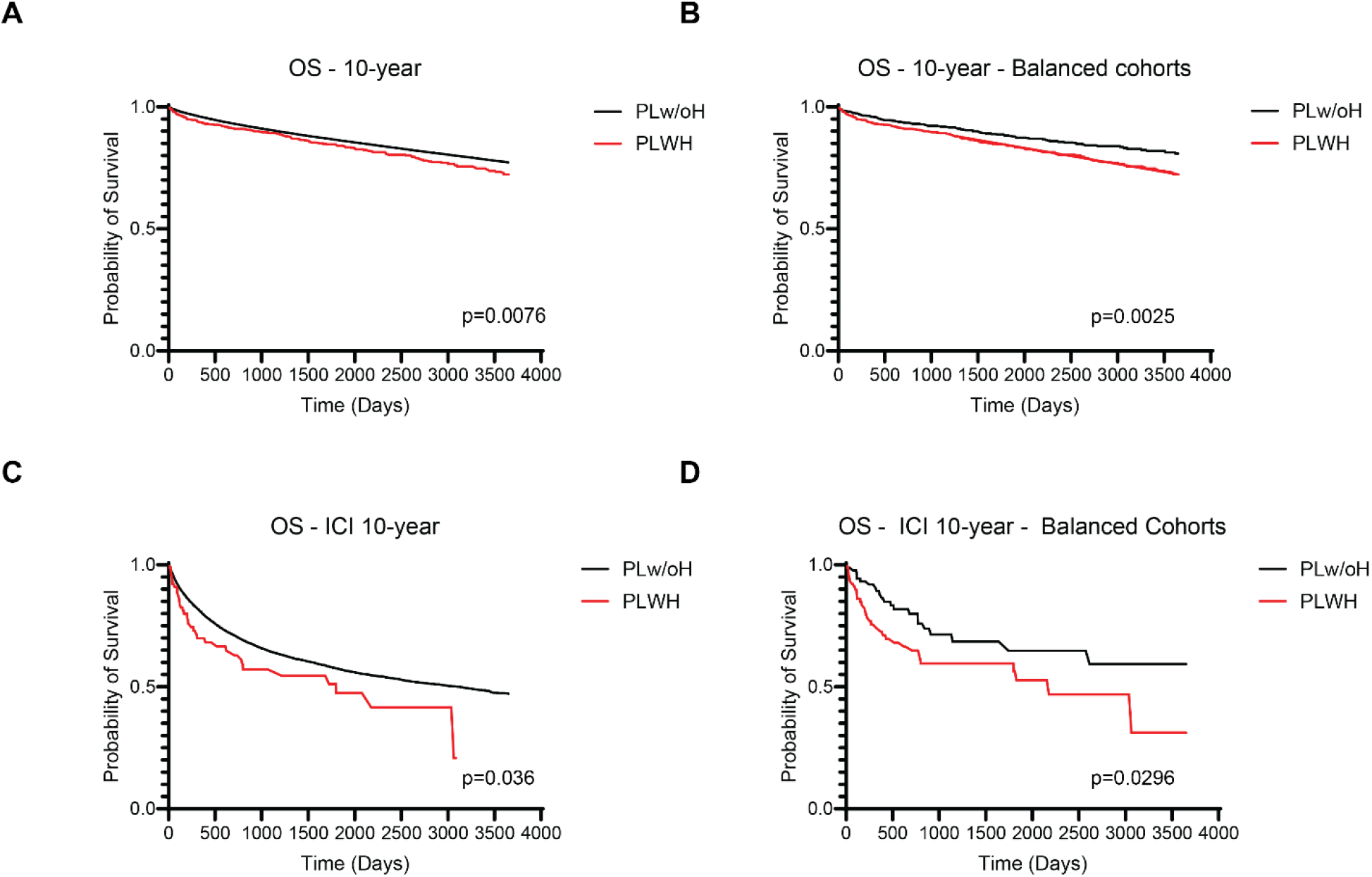}
    \par
  }
  \par\smallskip
  \textbf{Question:} In panel A, what is the probability of survival at the rightmost end of the PLWH curve?\\
  \textbf{Answer:} 0.7229
\end{examplebox}

\clearpage
\newpage

\subsection{Human Verification Protocol}
\label{app:human_eval_protocol}

Four reviewers manually reviewed all 1,205 candidate questions. Three reviewers each checked 300 questions, and the fourth checked the remaining 305. 
Reviewers used WebPlotDigitizer \citep{rohatgi2017webplotdigitizer} to recover the target value from the figure and then compared it with the generated answer. A candidate passed the initial check if the recovered value was within 2\% relative error of the generated answer. Candidates were flagged for
further review if the discrepancy exceeded 2\%, the target
could not be uniquely identified, or the figure could not be
read reliably.

A fifth reviewer adjudicated all 131 flagged candidates using
WebPlotDigitizer and by examining the reconstruction recipe
and underlying source data. Candidates were retained
unchanged when the original recipe and answer were confirmed
to be correct. When an error in the recipe could be resolved,
the recipe was corrected and the answer was recomputed
from the source data and checked for consistency with the figure.
Reasons for exclusion included unverifiable reconstruction recipes, figures in which the target value could not be read reliably, and disagreements between figures and their deposited source data. The 2\% threshold therefore triggered further review rather than automatic exclusion. This process yielded 1,119 verified questions.

\section{Additional Results}

\subsection{Accuracy by Chart Type}
\label{app:accuracy_chart_type}
\begin{figure}[htbp]
    \centering
    \includegraphics[width=0.8\textwidth]{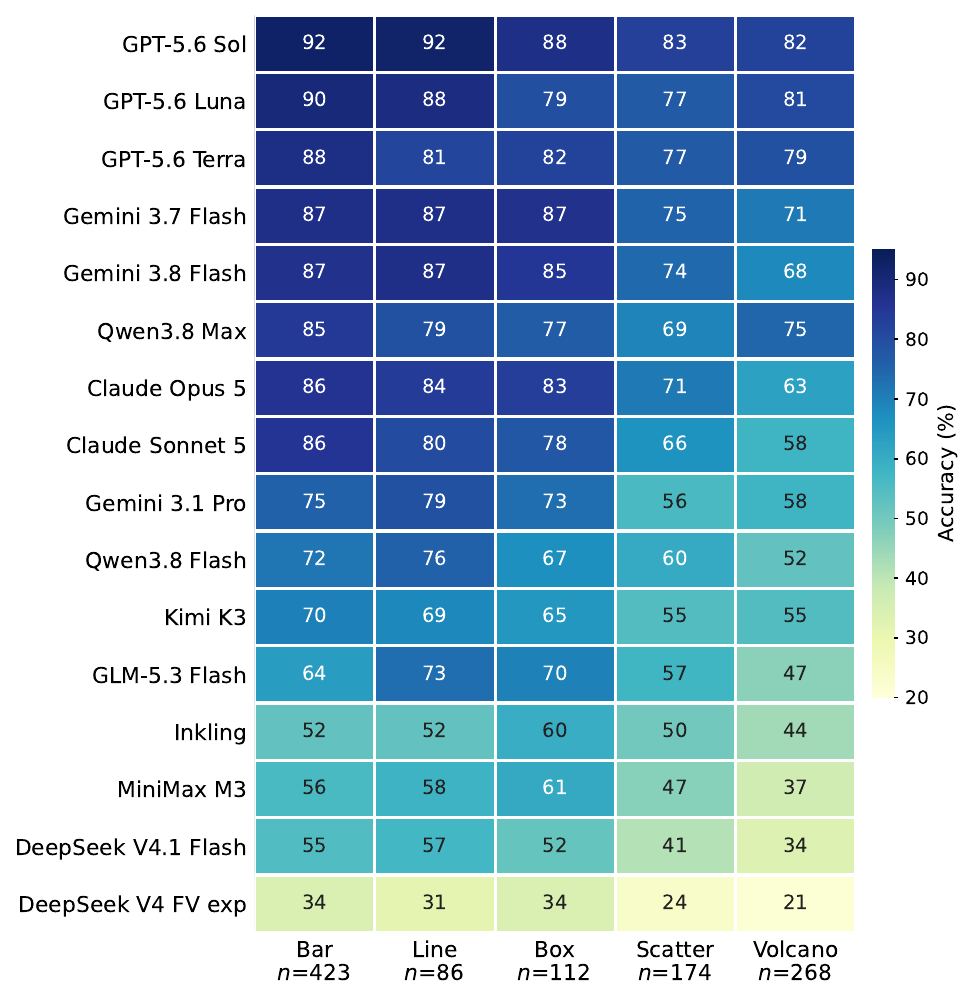}
\caption{
\textbf{Accuracy by chart type.} Per-model accuracy at a $\pm$5\% relative tolerance for the models in Table \ref{tab:results}. Columns show the five most common chart types, ordered by mean accuracy across models. Among these five types, either scatter or volcano plots yield the lowest accuracy for each of the 16 models. This may reflect the difficulty of locating a labelled target point among many others.
}
\label{fig:accuracy_by_type}
\end{figure}

\subsection{Figure--Source Data Discrepancy Example}
\begin{figure}[htbp]
    \centering
    \includegraphics[width=1.00\textwidth]{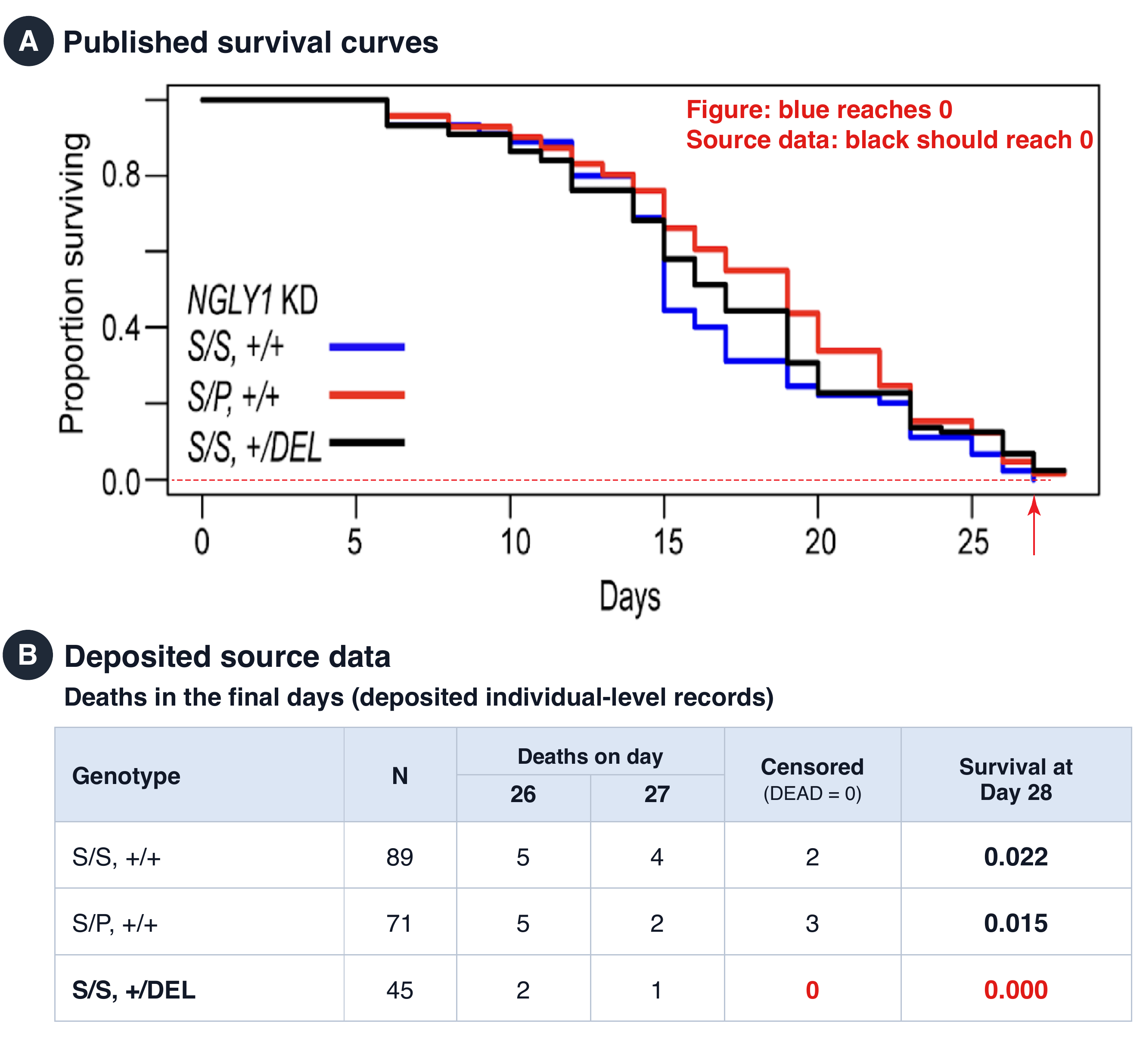}
\caption{
\textbf{Additional example of a discrepancy between a published figure and its deposited source data.}
The published survival curves differ from the values reconstructed from the deposited individual-level records. Although the discrepancy is visible throughout the trajectory, it is clearest at the endpoint: the published figure shows the blue curve ($S/S,+/+$) reaching zero survival first, whereas the deposited records imply zero survival for $S/S,+/DEL$. Survival at day 28 is the Kaplan--Meier estimate; two of the three censored $S/P,+/+$ animals were censored on day 21, so the estimate (0.015) differs from the simple proportion.
}
\label{fig:survival}
\end{figure}

\clearpage
\newpage
\section{Prompts used in the study}
\subsection{Plot Digitization Evaluation Prompt}
\label{app:prompt-eval}

Each model receives a single user message containing the figure image followed
by the question and a fixed answer-format suffix. No system prompt is used.
Figures are converted from TIF to PNG and downscaled so that the longest edge is
at most 1600\,px. The suffix constrains the output format only: it gives no
hint about how to read the figure (axes, ticks, interpolation) and explicitly
permits step-by-step reasoning, so that models without internal reasoning are
not forced to answer cold.

\begin{promptbox}
\textit{[figure image]}

\ph{question}

Work through the figure step by step, then end your reply with your final answer
as a single JSON object on its own line:\\
\texttt{\{"answer":\ <number, or null>\}}
\end{promptbox}

\subsection{Question Rewriting Prompt}
\label{app:prompt-rewrite}

\begin{promptbox}[Question Rewriting Prompt]
You are inside the directory of a bioRxiv preprint (full text XML and PDF, figure
images as TIF, and supplementary data files under \texttt{content/supplements/}).
The directory is read-only; write any scratch files to \texttt{/tmp}.

We have a benchmark question that asks for one numerical value plotted in a
figure of this paper. The question is currently phrased in terms of the drawing
(``panel'', ``bar'', ``point'', ``y-axis'', \ldots). Your job is to REWRITE it so
that it asks for the same value in the paper's own scientific terms, without
referring to the figure or to any table.

\textbf{\#\# The value}

\begin{tabular}{@{}lp{0.72\linewidth}@{}}
Original question: & \ph{question} \\
Figure file:       & \ph{figure\_file} (panel \ph{panel\_id}, chart type: \ph{chart\_type}) \\
Gold answer:       & \ph{gold} \ph{unit} \\
\end{tabular}

How the gold was computed from the deposited source data:\\
\ph{recipe}

\textbf{\#\# Rules for the rewritten question}

\begin{enumerate}[leftmargin=*, itemsep=0.3em, topsep=0pt]
  \item Do not mention figures, panels, plots, bars, points, lines, axes, colors,
        legends, labels, or anything about how the value is drawn. Do not mention
        tables, sheets, columns, rows, or supplementary files either.
  \item Describe the value the way the paper would in its text or methods: the
        experiment or analysis, the sample / group / condition / gene / term, the
        measured quantity, and the statistic (a single measurement, a mean, a
        median, a count, a fold change, \ldots). Use the terminology of the paper
        itself (main text, methods, figure caption). Read the paper to get this
        right.
  \item Do NOT copy internal identifiers that appear only in the source data file
        (sheet names, cryptic column headers such as \texttt{study\_count} or
        \texttt{padj}, sample codes) unless the paper text also uses them.
  \item The rewritten question must identify exactly ONE value in the paper, and
        it must be the gold value above. If the same quantity was measured under
        several conditions, time points, replicates, or normalizations, spell out
        which one. Check the paper and the data for such collisions and rule them
        out in the wording rather than with a figure reference.
  \item Keep the statistic identical to the original (if the original asks for a
        median, ask for the median; if it asks for a mean of replicates, ask for
        the mean). State the unit if the gold has one.
  \item One sentence, or two if needed for disambiguation. Ask for a single number.
\end{enumerate}

\textbf{\#\# Before answering}

Open the source data file and follow the recipe to confirm that it yields the
gold value; report the recomputed value. Read the figure caption and the
relevant text so the wording matches the paper.

Then check whether the paper ITSELF already states the answer: search the full
text XML (abstract, results, methods, figure captions, any in-text tables) for
the gold value, including rounded forms (e.g.\ \texttt{209.84 -> "210"},
\texttt{"\textasciitilde 2.1 x 10\textasciicircum 2"}), other units or scalings
(pg/mL vs ng/mL, fraction vs percent) and the value embedded in a range or a
``mean +/- sd''. If the text gives the value closely enough that a reader could
answer the question without the figure or the data (within a few percent), set
\texttt{value\_in\_text} to true and quote the sentence. This does not change
how you write the question; such questions are simply excluded downstream.

Return the result as a JSON object matching the required schema.
\end{promptbox}

\subsection{Evidence-Representation Agent Prompt}
\label{app:evidence-eval}

\begin{promptbox}
You are in the working directory of a bioRxiv preprint. It contains files from
the paper: the full text as XML and, depending on what is available, figure
images and/or supplementary data files. Use whatever is in the directory to
answer the question below about this paper.

Question: \ph{question}

Work through it step by step, then end your reply with your final answer as a
single JSON object on its own line:\\
\texttt{\{"answer":\ <number, or null>\}}
\end{promptbox}

\end{document}